\documentclass[letterpaper, 10 pt, conference]{ieeeconf}  

\IEEEoverridecommandlockouts                              

\usepackage{graphicx}%
\usepackage{multirow}%
\usepackage{amsmath,amssymb,amsfonts}%
\usepackage{mathrsfs}%
\usepackage{xcolor}%
\usepackage{textcomp}%
\usepackage{manyfoot}%
\usepackage{booktabs}%
\usepackage{algorithm}%
\usepackage{algorithmicx}%
\usepackage[noend]{algpseudocode}%
\usepackage{listings}%
\usepackage{caption}
\usepackage{subcaption}
\usepackage{svg}
\usepackage{stfloats}
\usepackage{float}
\usepackage{placeins}
\usepackage{cite}
\usepackage{hyperref}

\usepackage{url}

\algnewcommand{\algorithmicgoto}{\textbf{go to}}%
\algnewcommand{\Goto}[1]{\algorithmicgoto~\ref{#1}}%

\title{\LARGE \bf
Multi-Agent Path Finding Using Conflict-Based Search and Structural-Semantic Topometric Maps
}

\author{Scott Fredriksson$^{\dagger*}$, Yifan Bai$^\dagger$, Akshit Saradagi and George Nikolakopoulos%
\thanks{$^{\dagger}$The Authors contributed equally.}
\thanks{$^*$ Scott Fredriksson is the corresponding author (email : scofre@ltu.se). The authors are with the Robotics and AI group, in the Department of Computer Science, Electrical and Space Engineering at Luleå University of Technology, Sweden.}%
}
\begin{document}
\newcommand{\customcomment}[1]{ // #1}
\maketitle

\begin{abstract} As industries increasingly adopt large robotic fleets, there is a pressing need for computationally efficient, practical, and optimal conflict-free path planning for multiple robots. Conflict-Based Search (CBS) is a popular method for multi-agent path finding (MAPF) due to its completeness and optimality; however, it is often impractical for real-world applications, as it is computationally intensive to solve and relies on assumptions about agents and operating environments that are difficult to realize. 
This article proposes a solution to overcome computational challenges and practicality issues of CBS by utilizing structural-semantic topometric maps. Instead of running CBS over large grid-based maps, the proposed solution runs CBS over a sparse topometric map containing structural-semantic cells representing intersections, pathways, and dead ends. This approach significantly accelerates the MAPF process and reduces the number of conflict resolutions handled by CBS while operating in continuous time. 
In the proposed method, robots are assigned time ranges to move between topometric regions, departing from the traditional CBS assumption that a robot can move to any connected cell in a single time step. The approach is validated through real-world multi-robot path-finding experiments and benchmarking simulations. The results demonstrate that the proposed MAPF method can be applied to real-world non-holonomic robots and yields significant improvement in computational efficiency compared to traditional CBS methods while improving conflict detection and resolution in cases of corridor symmetries.  
\end{abstract}

\section{Introduction}
Multi-agent path-finding (MAPF)\cite{stern2019multi} is a fundamental problem in artificial intelligence and robotics, where the objective is to find collision-free paths for multiple agents, from their starting positions to their respective target positions within a shared environment, while minimizing the total path cost of all agents or a similar criteria. This problem is critical in domains such as warehouse automation \cite{ma2019lifelong}, robotics \cite{honig2018trajectory}, and video game AI \cite{stern2019multi}, where efficient and conflict-free coordination of multiple agents is essential.

There are two general approaches to MAPF. The first is distributed reactive solutions, where agents dynamically resolve conflicts as they appear during traversal toward the goal position \cite{ma2021distributed, lafmejani2021nonlinear, park2023dlsc, Multi_agent_Niklas}. The second approach is a centralized method \cite{standley2010finding}, where collision-free paths are calculated before the robots start moving toward their goals. In this article, the focus is on the latter approach. The centralized MAPF approaches are known to be NP-hard \cite{yu2015optimal}, presenting significant challenges due to the computational complexity involved. Over the years, various methods have been developed to tackle the centralized MAPF problem, including prioritized planning\cite{ma2019searching}, sub-dimensional expansion\cite{wagner2015subdimensional}, and conflict-based search (CBS)\cite{sharon2015conflict}. Among these, CBS has emerged as a popular approach due to its completeness and optimality\cite{sharon2015conflict}. CBS incrementally resolves conflicts between agents' paths, ensuring the resulting solution is feasible and optimal. Recently, CBS has been improved and enhanced\cite{boyarski2015icbs, gange2019lazy} and integrated with model predictive control\cite{via2020efficient, tajbakhsh2024conflict} and topological maps\cite{mcbeth2023scalable} for multi-robot motion planning.

Although CBS is optimal, it has two significant drawbacks with respect to computational speed and practicality. Computationally, 
CBS is an iterative approach wherein a path is calculated for one agent in each iteration and evaluated for new vertex and edge conflicts. As the number of cells and agents grows, CBS requires more iterations, leading to large computation times, especially in cases with pairwise symmetry. Pairwise symmetry occurs when two agents have multiple potential paths to their goals that, while individually promising, conflict when combined. This phenomenon manifests as rectangle symmetry, target symmetry, and corridor symmetry~\cite{li2019symmetry,li2021pairwise}.
In the case of practically, CBS makes assumptions that do not translate well to the real world. Firstly, it assumes that a robot is the size of a single grid cell, which is impractical in environments where large cells are infeasible or when robots have non-rectangular footprints. Secondly, it assumes that an agent can move to any neighboring cell in one time step, which is unrealistic, as few robots can perform such motions.

To address the issues of CBS regarding computational speed and practicality, this article proposes a novel solution wherein a structural-semantic topometric map is used instead of a grid-based map, to perform CBS. A topometric map is a hybrid approach that combines aspects of a topological map \cite{Remolina2004} with a grid-based metric map \cite{Kostavelis2015}. 
In this article, we utilize a structural-semantic topometric map generated using a method from our previous work \cite{FREDRIKSSON_SEMANTIC_MAPPING}, which extracts a topometric map with structural-semantic regions such as intersections, pathways, dead ends, and pathways leading to unexplored areas from a grid-based map.

The main contribution of this article is a novel methodology for MAPF that integrates structural-semantic topometric maps with Conflict-Based Search (CBS) and Enhanced Conflict-Based Search (ECBS), to overcome issues in traditional CBS approaches, such as corridor symmetry and slow path planning. The proposed method also improves the practicality of CBS and ECBS in real-world scenarios by taking traversal times of agent into account. To achieve this, the article: i) redefines the CBS and ECBS problem in continuous time and makes adaptations suitable for a topometric map; ii) generalizes the notions of edge and vertex conflicts and introduces a method for detecting such conflicts in continuous time for topometric cells; and iii) presents a path planning algorithm that computes paths on a topometric map in continuous time, while taking the time constraints associated with topometric cells into account.

The proposed method is validated in two stages.
Firstly, the proposed method is evaluated in simulations and compared to the standard CBS and ECBS methods to demonstrate its performance gains. Secondly, the proposed method was validated in real-world experiments with non-holonomic robots to demonstrate its practicality.

\section{Problem Description}
Consider a planar grid-world with width $W \in \mathbb{Z_+}$ and length $H \in \mathbb{Z_+}$. We define the set of locations on the grid $\mathcal{L}=\{0, \ldots, W-1\} \times\{0, \ldots, H-1\}$, where each location $l\in \mathcal{L}$ is represented by a pair of coordinates $l=(x, y)$, with $x$ and $y$ denoting the horizontal and vertical positions in the plane. Let $\mathcal{U} \subset \mathcal{L}$ denote the set of occupied locations, which are positions in the world that the agents cannot traverse or occupy. 

To represent the environment, the grid $\mathcal{L}$ is processed to construct a Topometric map $\mathcal{T}=(\mathcal{A},\mathcal{O})$ (using the algorithm proposed in \cite{FREDRIKSSON_SEMANTIC_MAPPING}), where $\mathcal{A}=\{A_1, \dots, A_K\}$ is the set of vertices of the topometric map and represents a set of semantically meaningful regions in the world which the agent can traverse, i.e., $\mathcal{A}\cap\mathcal{U}=\varnothing$. Two examples are shown in Figures \ref{fig:grid_map_PM} and \ref{fig:exMap}, where the grid-world has been segmented into structural-semantic regions such as intersections, pathways and dead-ends. The regions in $\mathcal{T}$ are connected by a set of openings $\mathcal{O}=\{O_1, \dots, O_T\}$ which forms the set of edges of the topometric map, connecting each region $A_k$ to its neighbors.

A Multi-agent Path Finding (MAPF) instance is defined on the topometric map $\mathcal{T}$ for a set of agents $R=\{a_1, \ldots, a_n\}$ (where $n$ is the number of agents), a tuple of distinct starts $R_{start}=(r_{start}^1, \ldots, r_{start}^{n})$ and goals $R_{goal}=(r_{goal}^1, \ldots, r_{goal}^{n})$, with $r_{start}^{i}$, $r_{goal}^{i} \in \mathcal{A}$ corresponding to the initial position and target position of agent $a_i$. Let $\pi_a^t \in \mathcal{L}$ denote the location of an agent $a$ at time $t \in \mathbb{R}_{+}$. A path $p_a= \left[\pi_a^{t_0}, \pi_a^{t_1}, \ldots, \pi_a^{T^a}\right]$ for agent $a$ is a sequence of locations such that $\pi_a^{t_0} = r_{start}^a$ and $\pi_a^{T_a} = r_{goal}^a$. For all $t>T_a$, $\pi_a^t=r_{goal}^a$, as the agent remains at its goal location after the path concludes. 

A solution of MAPF is a set of conflict-free paths $\left\{p_{1}, \ldots, p_{n}\right\}$ for the $n$ agents. 
More specifically, each agent $a$ must abide by a set of constraints $\mathcal{C}_a=\{C_{A_1},\dots, C_{A_K}\}$, where $C_{A_k}=\{c_{A_{k}1},\dots, c_{A_{k}L}\}$ is a set of time constraints for the region $A_k$ and $c_{A_{k}l}=[T_{start}^{A_{k}l}, T_{end}^{A_{k}l}]$ is a time range for which the region $A_k$ is restricted for agent $a$, as other agents are traversing through $A_k$.

The MAPF goal is to minimize the sum-of-costs $(S O C)$ metric of all agents: $\sum_{a_i \in A} T_i$, where $T_i$ is the total time agent $a_i$ takes to traverse its path and reach the goal $r_{goal}^a$, including the waiting times. 
\section{Methodology}
This section presents a novel methodology for MAPF, in which a conflict-based search (CBS) is performed over a structural-semantic topometric map. 
\subsection{Conflict-based Search} \label{sec:cbs}
In traditional CBS~\cite{sharon2015conflict}, two types of conflicts between agents are typically identified: vertex conflicts, where two agents occupy the same cell simultaneously, and edge conflicts, where two agents attempt to traverse the same edge between adjacent cells in opposite directions during the same time step. Constraints are then applied to prevent these conflicts by disallowing an agent from occupying a particular vertex or traversing an edge at a specific time.
\begin{figure}[tb]
    \centering
    \begin{subfigure}[b]{.4\linewidth}
         \centering
         \includegraphics[width=\linewidth]{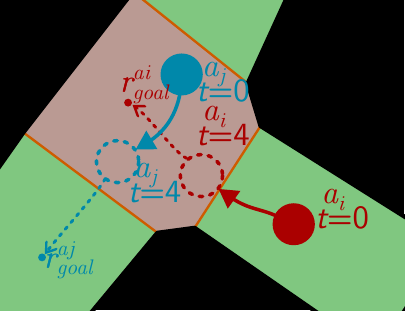}
         \caption{Region conflict}
         \label{fig:regionC}
    \end{subfigure}
    \begin{subfigure}[b]{.4\linewidth}
         \centering
         \includegraphics[width=\linewidth]{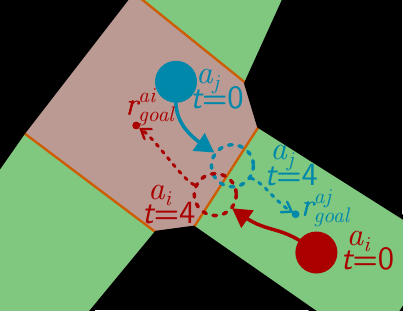}
         \caption{Opening conflict}
         \label{fig:openingC}
    \end{subfigure}
    \caption{Illustration of the region and opening conflicts in a structural semantic topometric map. In (a), $a_i$ enters a region already occupied by agent $a_j$. In (b), two agents are trying to go through the same opening at the same time.}
    \label{fig:conflict}
\end{figure}

When adopting CBS for use over topometric maps, several modifications are required to account for the unique structure of these maps. Instead of conflict detection at a single cell and a single time step, topometric maps necessitate conflict detection over larger areas and extended time ranges. 
This article presents definitions for conflicts over topometric maps, by suitably adapting the classical notions of vertex and edge conflicts. In this adaptation, conflicts are defined as either \textbf{region conflict}, denoted as $\mathcal{RC}=\left(a_i, a_j, A_k,[t_{\text {start }}^i, t_{\text {end }}^i],[t_{\text {start }}^j, t_{\text {end }}^j]\right)$, where agents $a_i$ and $a_j$ occupy the same area $A_k$ during overlapping time intervals i.e, $[t_{\text {start }}^i, t_{\text {end }}^i] \cap [t_{\text {start }}^j, t_{\text {end }}^j]\neq \emptyset$ ; or \textbf{opening conflict}, denoted as $\mathcal{OC}=\left(a_i, a_j,[A_k,A_l],[t_{\mathrm{start}}^i, t_{\text {end }}^i],[t_{\mathrm{start}}^j, t_{\text {end }}^j]\right)$, where agent $a_i$ moves from $A_k$ to $A_l$ and will occupy $A_l$ during the time $[t_{start}^i,t_{end}^i]$, while agent $a_j$ moves from $A_l$ to $A_k$ and occupies $A_k$ during the time $[t_{start}^j,t_{end}^j]$, with $t_{start}^i = t_{start}^j$. The two types of conflicts are illustrated in Figure \ref{fig:conflict}. 

Both types of conflicts create two new constraints, one for each agent involved. A region conflict between agents $a_i$ and $a_j$ at region $A_k$ results in the following constraints (restrictions for entering $A_k$):
\begin{equation}
\begin{matrix}
    \mathcal{RC}=\left(a_i, a_j, A_k,[t_{\text {start }}^i, t_{\text {end }}^i],[t_{\text {start }}^j, t_{\text {end }}^j]\right) \\
    \downarrow\\
    \mathrm{rc}_i=\left(a_i, c_{A_kl}=[t_{\text {start }}^j, t_{\text {end }}^j]\right) \text{ and}\\
    \mathrm{rc}_j=\left(a_j, c_{A_kl}=[t_{\text {start }}^i, t_{\text {end }}^i]\right).\\
\end{matrix}
\end{equation}

An opening conflict results in the following constraints: 
\begin{equation}
\begin{matrix}
    \mathcal{OC}=\left(a_i, a_j,[A_k,A_l],[t_{\mathrm{start}}^i, t_{\text {end }}^i],[t_{\mathrm{start}}^j, t_{\text {end }}^j]\right) \\
    \downarrow\\
    \mathrm{oc}_i=\left(a_i,c_{A_kl}=[t_{\mathrm{start}}^j-\Delta t, t_{\text {end }}^j]\right), \text{ and } \\\mathrm{oc}_j=\left(a_j,c_{A_ll}=[t_{\mathrm{start}}^i-\Delta t, t_{\text {end }}^i]\right)\\
\end{matrix}
\end{equation} 
where $\Delta t$ is a small duration of time.

\begin{algorithm}[tb]
\caption{High-level of CBS on a topometric map}
\label{alg:cbs}
\begin{algorithmic}[1]
\Require Topometric map, start/goal areas of agents
\State $Root.constraints \leftarrow \emptyset$ \label{line:start1}
\State $Root.paths \leftarrow$ find individual paths using the low-level()
\State $Root.cost \leftarrow SOC(Root.paths)$
\State insert $Root$ to OPEN \label{line:start2}
\While{OPEN not empty} \label{line:bfs1}
    \State $P \leftarrow$ best node from OPEN \customcomment{lowest cost} \label{line:bfs2}
    \small
    \State Validate paths $\left\{p_1, \ldots, p_n\right\}$ in $P$ until a conflict occurs. \label{lines:return1}
    \normalsize
    \If{$P$ has no conflict} 
        \State \Return $P.paths$ \customcomment{$P$ is goal}
    \EndIf \label{lines:return2}
    \State $\mathcal{RC}/\mathcal{OC}\leftarrow$ first conflict in $P$ \label{line:detect conflict} 
    \For{each agent $a_k$ in $\{a_i, a_j\}$}\label{line:newnode_start}
        \State $A \leftarrow$ new node 
        \small
        \State $A.constraints \leftarrow P.constraints+ \mathrm{rc}_k/\mathrm{oc}_k$\label{line:newnode_end}
        \normalsize
        \State $A.paths \leftarrow P.paths$\label{line:lowlevel_start}
        \State $\mathcal{C}_{a_k} \leftarrow$ get constraints for $a_k$ in $A.constraints$
        \State Update $A.paths$ by invoking low-level ($a_k$,$\mathcal{C}_{a_k}$)
        \State $A.cost \leftarrow SOC(A.paths)$
        \State Insert $A$ to OPEN \label{line:lowlevel_end}
    \EndFor
\EndWhile
\end{algorithmic}
\end{algorithm}
Algorithm \ref{alg:cbs} shows the overall search structure of CBS to find conflict-free paths. The CBS operates on a best-first search [Lines~\ref{line:bfs1}-\ref{line:bfs2}] of a binary tree structure called a Constraint Tree (CT). Each node in the CT, referred to as a CT node, contains three elements: (1) a set of constraints that prohibit each agent $a_i$ from being in a specific area $A_k$ during the time range $c_{A_{k}l} = [T_{start}^{A_{k}l}, T_{end}^{A_{k}l}]$; (2) a set of paths $p_a$ for each agent that adhere to these constraints; and (3) a cost, which is the sum of the total time of traversing all paths.

CBS begins with a root node that has no constraints and independently plans a path for each agent [Lines\ref{line:start1}-\ref{line:start2}]. From there, CBS incrementally expands the CT by identifying conflicts between agents' paths. The algorithm follows a two-level approach: the high-level search and the low-level search.

The high-level search detects conflicts (either a region conflict $\mathcal{RC}$ or an opening conflict $\mathcal{OC}$) in the current node’s set of paths at the earliest possible time step. Conflict resolution involves creating two successor nodes in the CT. Besides inheriting the constraint of the parent node, each successor introduces an additional constraint (a region constraint $\mathrm{rc}_k$ or an opening constraint $\mathrm{oc}_k$) for one of the conflicting agents [Lines~\ref{line:newnode_start}-\ref{line:newnode_end}], effectively ensuring the conflict is avoided. The low-level search, which is described in subsection \ref{sec:pathPlan}, recalculates the path for each agent that respect the new set of constraints [Lines~\ref{line:lowlevel_start}-\ref{line:lowlevel_end}]. The algorithm continues this process of conflict detection and resolution until a solution is found where no conflicts remain, resulting in a set of collision-free paths $p_a$ [Lines~\ref{lines:return1}-\ref{lines:return2}].

\subsection{Bounded Sub-optimal Extension of CBS} \label{subsec:ECBS}
To enhance the performance of CBS (Conflict-Based Search) on topometric maps, we propose a bounded-sub-optimal extension by incorporating elements of Enhanced Conflict-Based Search (ECBS) \cite{barer2014suboptimal}. This modification trades a degree of optimality for improved computational speed, making it more practical for complex environments.

In standard CBS, the high-level search explores nodes in the constraint tree (CT) using best-first search, always expanding the node with the lowest cost. In our bounded sub-optimal extension, we substitute this best-first search with focal search, as done in ECBS. Focal search broadens the search by allowing the expansion of nodes that may not have the lowest cost but are within a suboptimality bound, defined by a weight $\omega$. This approach prioritizes nodes with fewer conflicts within the allowable cost range, enabling faster conflict resolution while maintaining a balance between optimality and efficiency.
Given that the topometric map divides the grid into structural-semantic regions for more efficient pathfinding, we retain the original best-first search in the low-level planner to maintain simplicity.

\subsection{Path Planning with Constraints}\label{sec:pathPlan}
Path planning with constraints, also referred to as low-level planning in CBS, finds a path for each agent $a$, one at a time, given a set of constraints $\mathcal{C}_a$ for the agent. Thus,  when discussing paths, constraints, start, and goal positions in this section, it is with respect to a single agent.

\subsubsection{Difference with respect to Traditional CBS}
In traditional CBS, the agent can take two actions each time step: move to a neighboring node or wait. When operating on a topometric map, an additional action is available: traveling within a semantic region (a node of a topometric map). The agent needs to physically traverse to the opening of the topometric region before it can move into a neighboring region on the topometric map. This, combined with the fact that the constraints for areas ($C_{A_k}$) are defined as time ranges instead of specific discrete time slots, as stated in subsection \ref{sec:cbs}, makes the traditional CBS approach for solving navigation with constraints impractical.
This is solved by using a search approach combining all three actions of an agent. In this approach, the agent will wait outside the region until all the constraints associated with the region are cleared. Upon entering the area, it will travel to the next opening. This process is repeated until the agent reaches its goal $r_{goal}^a$.

\subsubsection{Time Slots}
Each area $A_k$ has a set of constraints $C_{A_k} = \{c_{A_{k}1}, \dots, c_{{A_k}L}\}$, where each constraint is given by $c_{A_{k}l} = [T_{start}^{A_{k}l}, T_{end}^{A_{k}l}]$. From these constraints, available time slots can be extracted as $S_{A_k} = (0,\infty)\setminus C_{A_k} = \{s_{A_k1}, \dots, s_{A_kN}\}$, where each time slot is represented by $s_{A_{k}n} = (T_{start}^{A_{k}n}, T_{end}^{A_{k}n})$. There is always be one more time slot in $S_{A_k}$ than the number of constraints in $C_{A_k}$, i.e., $N=L + 1$. For example, a region without any constraints will have the available time slot $s_{A_{k}1} = (0, \infty)$, while an area with a single constraint will have the available time slots $s_{A_{k}1} = (0, T_{start}^{{A_k1}})$ and $s_{A_{k}2} = (T_{end}^{{A_k1}}, \infty)$.

\subsubsection{The Navigation Tree}
In this article, the A* method is used for path planning, with time as the heuristic. The search tree consists of two sets: an open set $\mathcal{N}_o$, which contains the actively searched nodes $n_i$, and a closed set $\mathcal{N}_c$, which contains the parent nodes  of those in $\mathcal{N}_o$. 

Each node in the open set $n_i \in \mathcal{N}_o$ contains the following information: a) The previous region $A_p^{n_i}$. b) The current region $A_c^{n_i}$. c) The next region $A_n^{n_i}$. d) A time-slot $s_{n_i}=(T_{current}^{n_i}, T_{end}^{n_i})$ where $T_{current}^{n_i}$ is the time at which an agent reaches the node's position in the world, which is the opening between $A_p^{n_i}$ and $A_c^{n_i}$. The agent can wait for constraints at the node's position until $T_{end}^{n_i}$. The value for $T_{current}^{n_i}$ is calculated using the following equation:
\begin{equation}
\label{eq:updateCurrent}
    T_{current}^{n_i} = \frac{|p_{n_i}|}{r_{speed}} * I_{margin}
\end{equation} 
where $|p_{n_i}|$ is the length of the combined path of the node $n_i$ and its parents, starting at the initial position $r_{start}^a$ of an agent and ending at the position of node $n_i$. The value $r_{speed}$ represents the speed at which the agents travel. Since this time estimation does not account for the time required for the agent to make turns or any stochastic variations in navigation, the time is scaled by the user-tunable variable $I_{margin}\in \mathbb{R}_{>1}$ to ensure the agent has enough time to reach each region in the topometric map.

\subsubsection{Expansion of the Navigation Tree}
Each time the A* search tree is expanded, a node that minimizes the cost in the equation below is selected from $\mathcal{N}_o$. 
\begin{equation}
\label{eq:next_node}
    \bar{n}_i=\arg \min_{n_i \in \mathcal{N}_o} \left( T_{current}^{n_i} + \frac{D(n_i,r_{goal}^a)}{r_{speed}} * I_{margin}\right) 
\end{equation}
where $D(n_i,r_{goal}^a)$ is the Euclidean distance from the node $n_i$ to the goal position $r_{goal}^a$.
Based on $\bar{n}_i$, its children are generated and added to $\mathcal{N}_o$, while $\bar{n}_i$ is then moved to the closed set $\mathcal{N}_c$.

\begin{algorithm}[h]
\caption{Expanding the navigation tree nodes in the Low-level planer of CBS}
\label{alg:path}
\begin{algorithmic}[1]
\Require Node $\bar{n}_i$ given by Eq. \eqref{eq:next_node} 
\For{each $(A_a,(T_{start}^{A_c^{\bar{n}_{i}}n},T_{end}^{A_c^{\bar{n}_{i}}n}))\in N(A_{n}^{\bar{n}_i}) \times S_{A_{c}^{\bar{n}_i}}$} \label{line:pathSearch}
    \If{$T_{start}^{A_c^{\bar{n}_{i}}n} \geq T_{end}^{\bar{n}_{i}}$} \label{line:con1}
        \State \Goto{line:pathSearch}
    \EndIf
    \If{$T_{end}^{A_c^{\bar{n}_{i}}n} \leq T_{current}^{\bar{n}_{i}}+ p_{time}^{\bar{n}_i}$} \label{line:con2}
        \State \Goto{line:pathSearch}
    \EndIf
    \If{$A_a=A_c^{\bar{n}_{i}}$ and $T_{end}^{A_c^{\bar{n}_{i}}n} \neq \infty$} \label{line:con3}
        \State \customcomment{Stop repeated moves}
        \State \Goto{line:pathSearch}
    \EndIf
    \State Create new child node $n_j$
    \State $A_p^{n_{j}} \leftarrow A_c^{\bar{n}_{i}}$
    \State $A_c^{n_{j}} \leftarrow A_n^{\bar{n}_{i}}$
    \State $A_n^{n_{j}} \leftarrow A_a$
    \State Update $T_{current}^{n_{j}}$ using Eq. \eqref{eq:updateCurrent}
    \State $T_{end}^{n_{j}} \leftarrow T_{end}^{A_c^{\bar{n}_{i}}n}$
    \State Add $n_{j}$ to $\mathcal{N}_o$
\EndFor
\State Move $\bar{n}_{i}$ to $\mathcal{N}_c$
\end{algorithmic}
\end{algorithm}
When a node $\bar{n}_i$ is found, Algorithm \ref{alg:path} is used to find and create child nodes $n_{j}$ for each time slot in the current region of node $\bar{n}_i$, i.e, $A_c^{\bar{n}_{i}}$ and for each region that is connected to $A_c^{\bar{n}_i}$ (collected into the set $N(A_{n}^{\bar{n}_i})$,). For a child node to be added for a particular time slot, the time slot and the parent node must fulfill three conditions, as shown in Algorithm \ref{alg:path}: a) On Line \ref{line:con1}: The start time of an available time slot $T_{start}^{A_c^{\bar{n}_{i}}n}$ must not start before the end time $T_{end}^{\bar{n}_{i}}$ of the node.
b) On Line \ref{line:con2}: The available time slot must not have already ended, and the node must have enough time to traverse the current region before the time slot ends. c) On Line \ref{line:con3}: If the new node $A_a$ is in the same region as the parent node, meaning that the child node will return to a previously visited node, it is only allowed to do so if there is a future constraint (i.e., the current time slot limit is not $\infty$). This prevents the search from getting stuck between two regions.
In addition to the last condition, a node whose current region is the same as its parent's previous region cannot take the first available time slot.

The tree expands until there is a node whose previous region is the same as the region where the goal position ( $R_{goal}^a$) is located. The complete path, $p_a$, is generated from the node and its parents and is returned to the high-level planner of CBS. 
\section{Validation}
This section presents results from the simulation and experimental validation of the MAPF approach proposed in this article. The computations were performed on a 1.9GHz AMD Ryzen 7 Pro 5850U laptop with 32GB RAM, running Linux kernel 6.10.9.
\subsection{Results from a benchmarking simulation setup}
\begin{figure}[b]
    \centering
    \begin{subfigure}[t]{.4\linewidth}
         \centering
         \includegraphics[width=\linewidth]{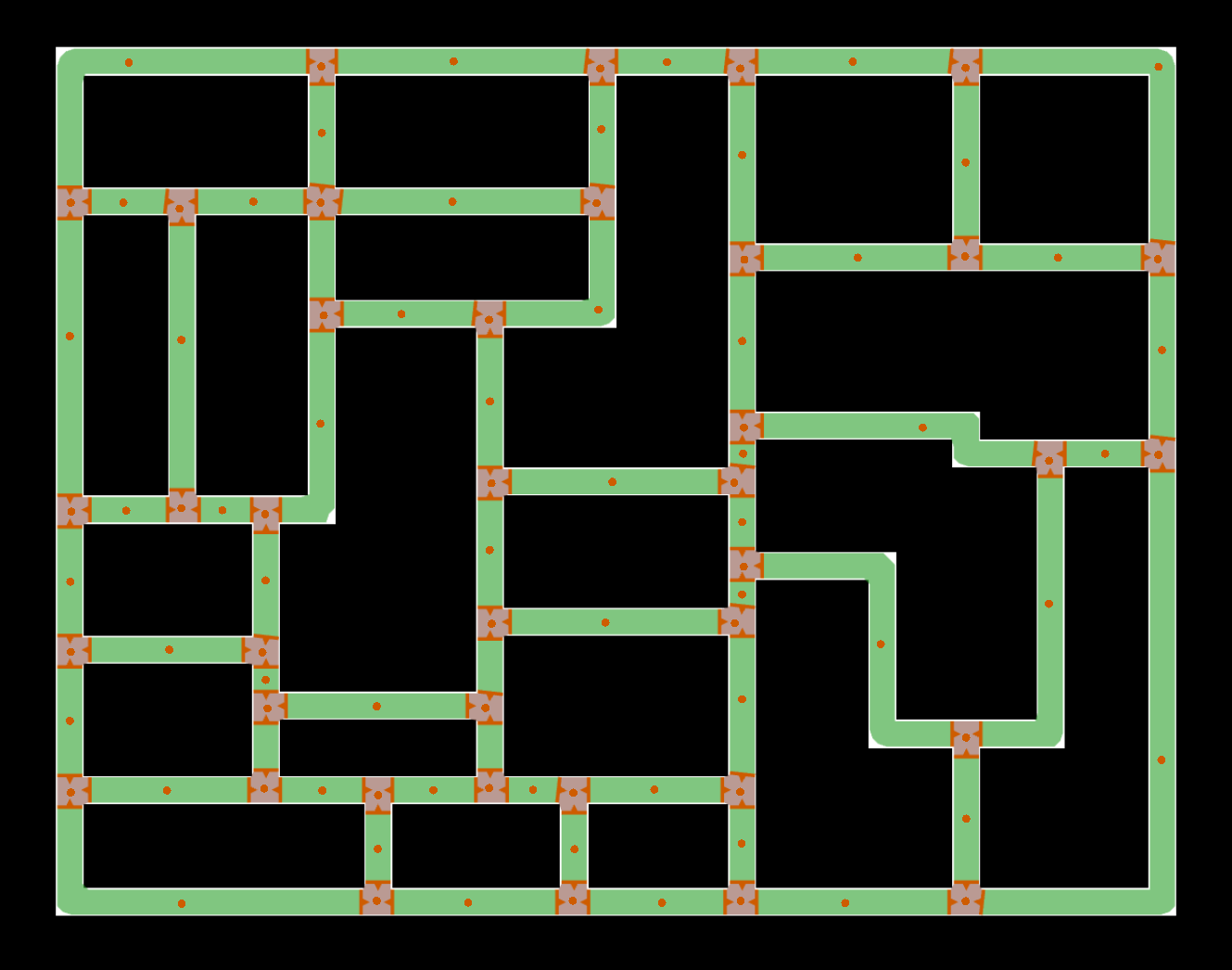}
         \caption{Map used for PM-CBS \& PM-ECBS}
         \label{fig:grid_map_PM}
    \end{subfigure}
    \begin{subfigure}[t]{.4\linewidth}
         \centering
         \includegraphics[width=\linewidth]{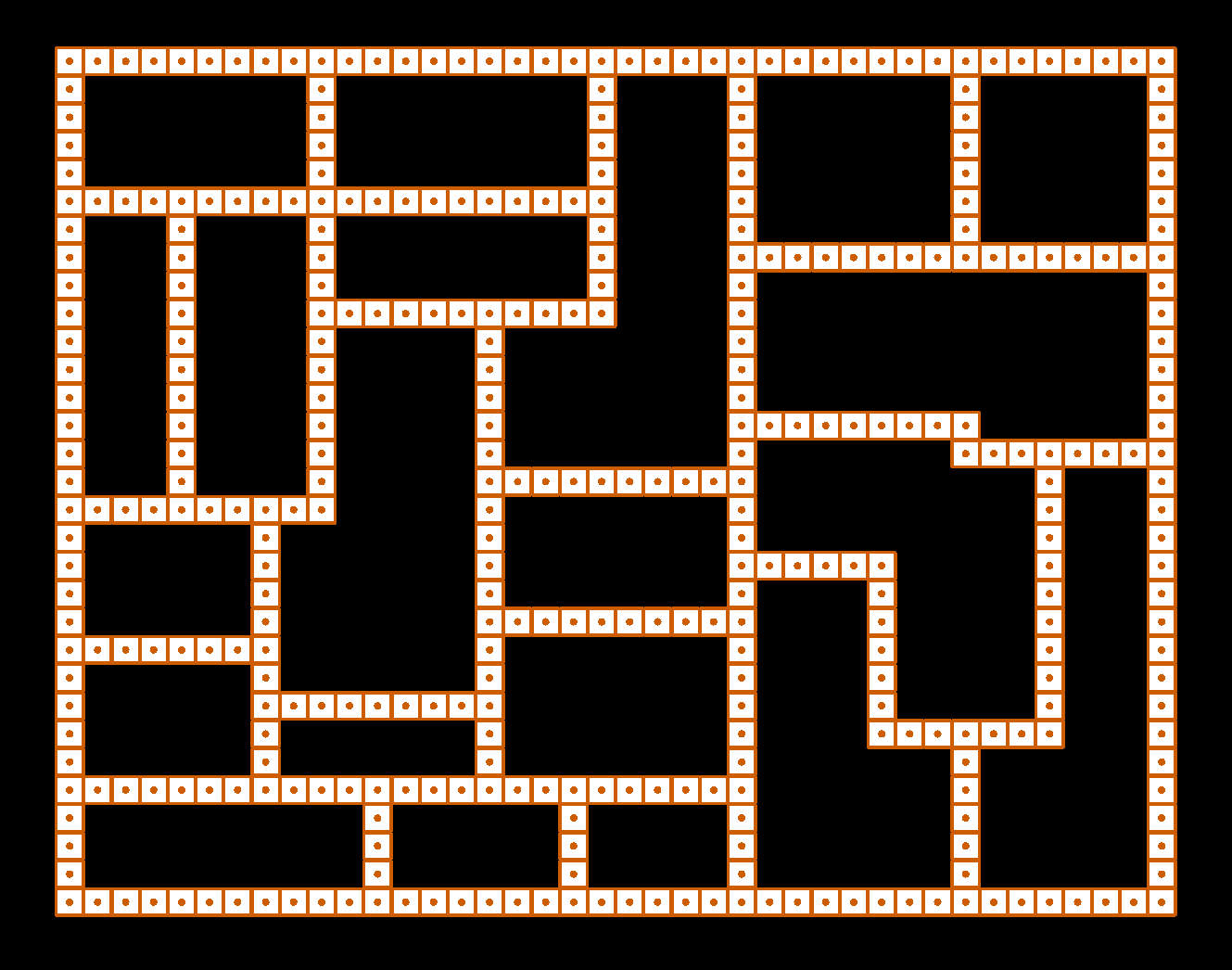}
         \caption{Map used for CBS \& ECBS}
         \label{fig:grid_map_CBS}
    \end{subfigure}
    \caption{Maps used for the comparative study. Vertices/regions \& edges/openings are marked with dots \& lines respectively.}
    \label{fig:gridmap}
\end{figure}
\begin{table}[b]
    \centering
    \resizebox{.9\linewidth}{!}{%
    \begin{tabular}{|c|cccc|} \hline
         Method & PM-CBS & CBS & PM-ECBS & ECBS \\ \hline
         
         Number of Agents & \multicolumn{4}{c|}{4}   \\ 
         
         Median Time (ms) & 1.09 & 3.35 & 0.58 & 3.85 \\

         Average Distance (Cells) & 100.44 & 103.05 & 101.24 & 103.10\\
         
         Success Rate (\%) & 100.00 & 98.80 & 100.00 & 100.00\\

         Median Expanded Nodes & 2 & 2 & 2 & 1  \\ \hline

         
         Number of Agents &  \multicolumn{4}{c|}{6}  \\ 
         
         Median Time (ms) & 1.86 & 23.49 & 0.97 & 9.99 \\

         Average Distance (Cells) & 152.67 & 156.21 & 154.22 & 157.29 \\
         
         Success Rate (\%) & 100.00 & 90.00 & 100.00 & 100.00 \\

         Median Expanded Nodes & 7 & 8 & 4 & 2  \\ \hline

         
         Number of Agents & \multicolumn{4}{c|}{8}  \\ 
         
         Median Time (ms) & 4.40 & 128.7 & 2.04 & 22.10 \\

         Average Distance (Cells) & 202.26 & 203.93 & 205.67 & 209.18  \\
         
         Success Rate (\%) & 98.40 & 68.20 & 100.00 & 99.80 \\

         Median Expanded Nodes & 27 & 42 & 10 & 3  \\ \hline
         
         
         Number of Agents   & \multicolumn{4}{c|}{10}  \\ 
         
         Median Time (ms) & 46.49 & 562.19 & 6.02 & 45.48 \\

         Average Distance (Cells) & 256.38 & 258.23 & 265.63 & 268.58 \\
         
         Success Rate (\%) & 88.80 & 35.40 & 97.60 & 98.60 \\

         Median Expanded Nodes & 913 & 285 & 50 & 5  \\ 
         \hline
    \end{tabular}}
    \caption{Results from the simulation-based benchmarking of the proposed method over the map shown in Figure \ref{fig:gridmap}.}
    \label{tab:sim_result}
\end{table}
To validate the performance of the proposed method, we compare it with traditional CBS \cite{sharon2015conflict} and Enhanced CBS (ECBS) \cite{barer2014suboptimal}. The proposed method, which integrates strucutral semantic-topometric maps with both CBS and ECBS to solve the Conflict Tree (CT), results in PM-CBS and PM-ECBS, respectively. PM-ECBS and ECBS are bounded sub-optimal variants of PM-CBS and CBS that trade off optimality for efficiency. They employ focal search at the high level and, in the case of ECBS, at the low level. This ensures that the solution found is within a sub-optimality factor $\omega$ of the optimal solution. In the simulations, $\omega = 1.2$ was used.

To evaluate the benefits of PM-CBS and PM-ECBS over CBS and ECBS, a $44 \times 38$ grid map with 385 cells of free corridor-like space was generated, as shown in Figure~\ref{fig:grid_map_CBS}. The vertices and edges of the grid map used by CBS and ECBS are also shown in Figure~\ref{fig:grid_map_CBS}. For the proposed PM-CBS and PM-ECBS methods, a topometric map with 98 regions was generated using the segmentation method described in \cite{FREDRIKSSON_SEMANTIC_MAPPING}, as shown in Figure~\ref{fig:grid_map_PM}, where each region and its corresponding openings are marked.

All the four methods under consideration were simulated using 4, 6, 8, and 10 agents to evaluate the success rate, runtime, and average combined distance of all agents in 500 instances, with a time limit of 30 seconds. For each instance, start and goal locations were randomly generated within regions of the topometric map. The start and goal regions were generated such that no start region was identical to another start region, and no goal region was identical to another goal region in the topometric map. All algorithms were tested under the same sets of start and goal locations.

\begin{figure}[t]
    \centering
    \includegraphics[width=.8\linewidth]{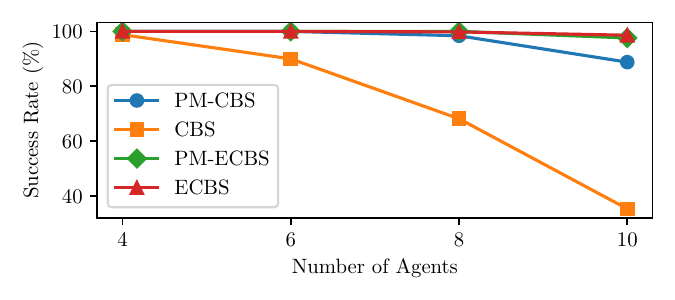}
    \caption{Success rates in conflict-free multi-agent path finding of the proposed methods and traditional CBS approaches.}
    \label{fig:success}
\end{figure}
Figure~\ref{fig:success} compares the success rates of the four algorithms as the number of agents increases. PM-CBS demonstrates a higher success rate than CBS, maintaining a 100\% success rate with six agents. However, as the number of agents increases, its performance gradually declines, reaching 88.8\% with ten agents. In contrast, CBS shows a sharp drop in success rate, falling to 35.4\% with ten agents, highlighting its limitation in handling maps with long corridors. On the other hand, PM-ECBS exhibits a similar success rate to that of ECBS. At ten agents, PM-ECBS achieves a success rate of 97.6\%, compared to 98.6\% for ECBS.

\begin{figure}[bt]
    \centering
    \begin{subfigure}[b]{.45\linewidth}
         \centering
         \includegraphics[width=\linewidth]{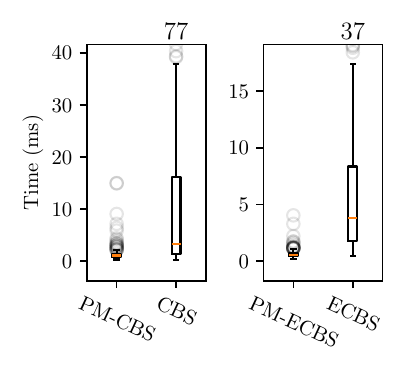}
         \caption{4 Agents}
         \label{fig:timeR4}
     \end{subfigure}
     \begin{subfigure}[b]{.45\linewidth}
         \centering
         \includegraphics[width=\linewidth]{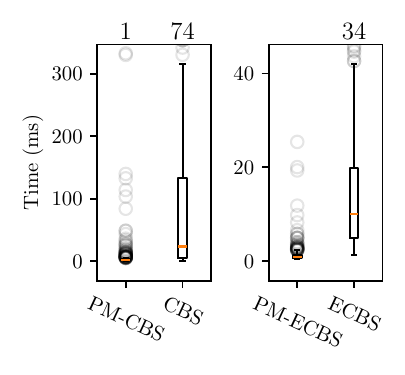}
         \caption{6 Agents}
         \label{fig:timeR6}
     \end{subfigure}
     \begin{subfigure}[b]{.45\linewidth}
         \centering
         \includegraphics[width=\linewidth]{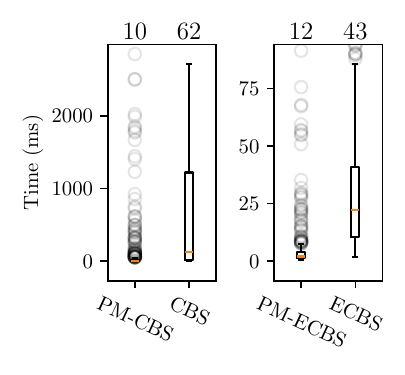}
         \caption{8 Agents}
         \label{fig:timeR8}
     \end{subfigure}
     \begin{subfigure}[b]{.45\linewidth}
         \centering
         \includegraphics[width=\linewidth]{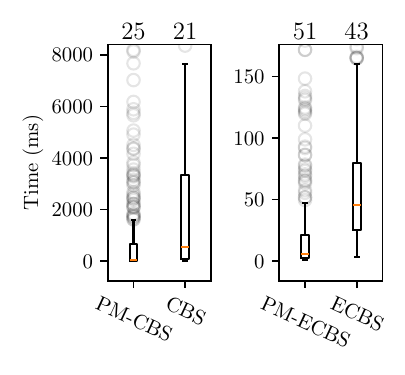}
         \caption{10 Agents}
         \label{fig:timeR10}
     \end{subfigure}
    \caption{Scaling of the computation time required by PM-CBS, PM-ECBS, CBS, and ECBS approaches with respect to the number of agents. The number of outliers that are outside the plot range is indicated on the top of each boxplot.}
    \label{fig:timeR}
\end{figure}
Figure~\ref{fig:timeR} presents boxplots comparing the runtime for the four algorithms across the four different agent counts (4, 6, 8, and 10), represented in subfigures~\ref{fig:timeR4}, ~\ref{fig:timeR6}, ~\ref{fig:timeR8}, and ~\ref{fig:timeR10}, respectively. Times from unsuccessful instances (when the time exceeds 30 seconds) are ignored in the boxplots. For all agent counts except ten, both PM-CBS and PM-ECBS exhibit a lower median runtime and less variation than CBS and ECBS, indicating superior performance under these conditions. Specifically, as shown in Table~\ref{tab:sim_result}, PM-CBS and PM-ECBS achieve a median runtime of 1.04 ms and 0.58 ms with four agents, compared to 3.35 ms for CBS and 3.85 ms for ECBS. 
However, as the number of agents increases to ten, there is a significant increase in the runtime of PM-CBS, which rises to 46.49 ms, slightly higher than ECBS at 45.48 ms but significantly lower than CBS at 562.19 ms. It is worth noting that PM-ECBS outperforms all the other solutions for ten agents, requiring a median time of only 6.02 ms.

As shown in Table \ref{tab:sim_result}, all agents' average combined distance is similar for all methods, meaning that they produce paths of the same length on average. However, as the number of agents increases, the path lengths produced by PM-ECBS and ECBS increase slightly compared to those produced by PM-CBS and CBS, which is to be expected since they prioritize solution speed over optimality.

The data in Table \ref{tab:sim_result} shows that PM-CBS and PM-ECBS are significantly more computationally efficient than their counterparts. The success rate for PM-CBS is significantly higher than that of CBS, while PM-ECBS has a similar success rate to that of ECBS. However, as seen in Table \ref{tab:sim_result}, when the number of agents increases, the number of expanded nodes (the depth of the search in the CT) grows significantly for PM-CBS and PM-ECBS. This is likely because the restrictions used by the proposed methods are more limiting than those of CBS and ECBS. Specifically, only one agent is allowed in each region in the proposed methods, whereas the traditional methods allow one agent in each cell within a region. As a result, there are fewer available solutions to the MAPF problem for the proposed methods, especially as the number of agents grows. Although median expanded nodes are higher for ten agents, the proposed method outperforms the corresponding methods as they are significantly more computationally efficient than the traditional methods.

\subsection{Experimental validation of the proposed MAPF approach}
\begin{figure}[t]
    \centering
    \begin{subfigure}[t]{.45\linewidth}
         \centering
         \includegraphics[width=\linewidth]{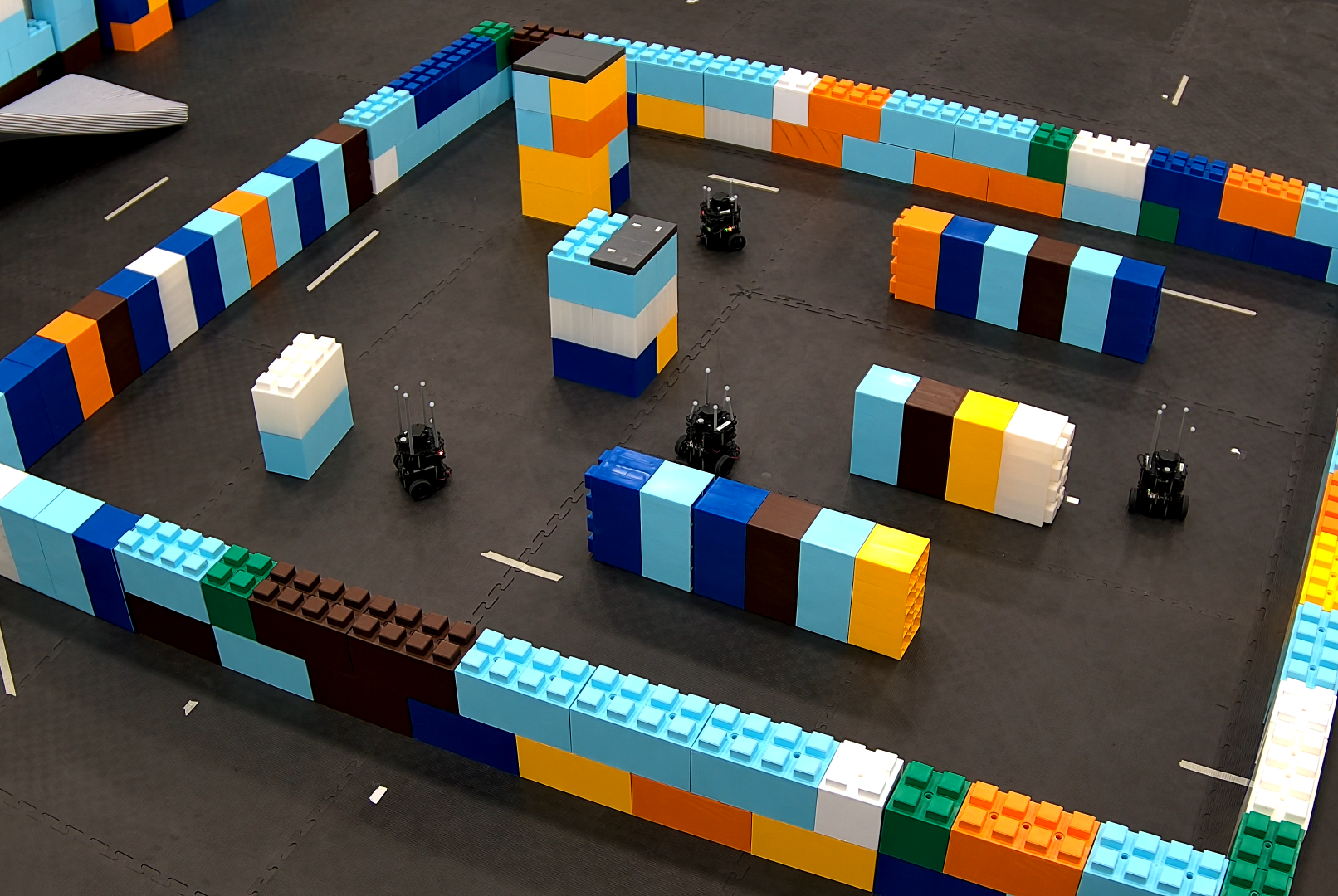}
         \caption{Experimental setup}
         \label{fig:exSetup}
    \end{subfigure}
    \begin{subfigure}[t]{.35\linewidth}
         \centering
         \includegraphics[width=\linewidth]{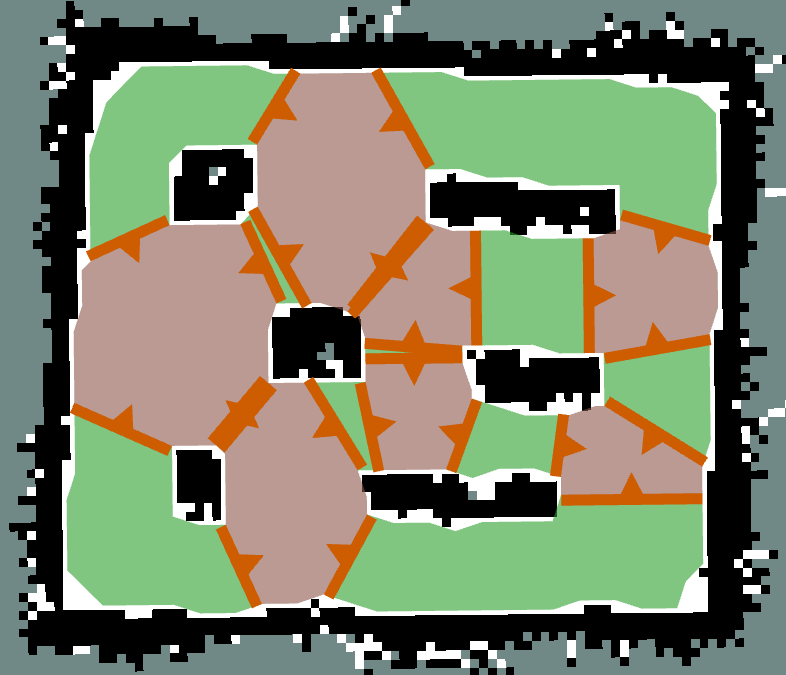}
         \caption{Semantic Topometric map}
         \label{fig:exMap}
    \end{subfigure}
    \caption{The experimental validation of the proposed method is performed in a test area, with Turtlebots as the agents.}
    \label{fig:expriment}
\end{figure}

The proposed PM-CBS method was validated in a real-world experiment to reinforce the claim that the proposed method improves the practicality of using  CBS. Four TurtleBot3 robots, which are small two-wheeled non-holonomic agents, were used in this experiment. A  test area with the size $3.25 \times 3.75$ m was constructed, as shown in Figure \ref{fig:exSetup}, and a 2D grid map of the test area was generated using the Hector SLAM \cite{Hector2011} method. Without manual modifications or cleanup, a topometric map with 19 regions was created following the method in \cite{FREDRIKSSON_SEMANTIC_MAPPING}. The 2D grid-based map and topometric map can be seen in Figure \ref{fig:exMap}.

\begin{figure}[t]
    \centering
    \begin{subfigure}[b]{.25\linewidth}
         \centering
         \includegraphics[width=\linewidth]{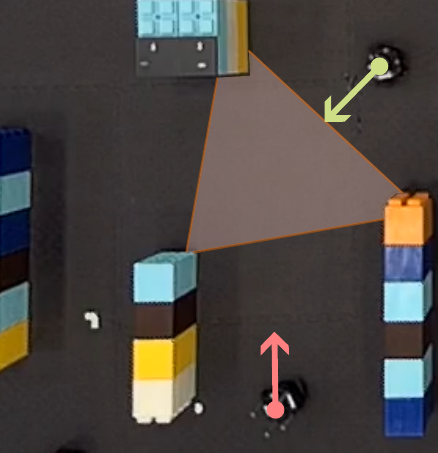}
         \caption{}
         \label{fig:expS1}
     \end{subfigure}
     \begin{subfigure}[b]{.25\linewidth}
         \centering
         \includegraphics[width=\linewidth]{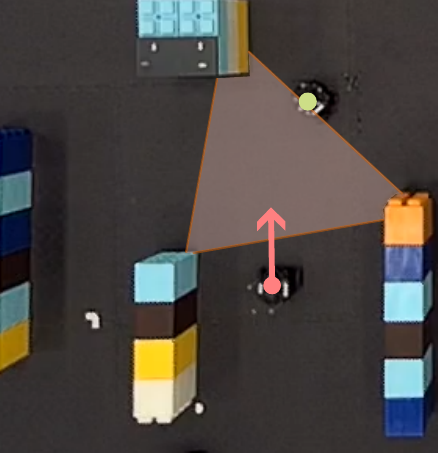}
         \caption{}
         \label{fig:expS2}
     \end{subfigure}
     \begin{subfigure}[b]{.25\linewidth}
         \centering
         \includegraphics[width=\linewidth]{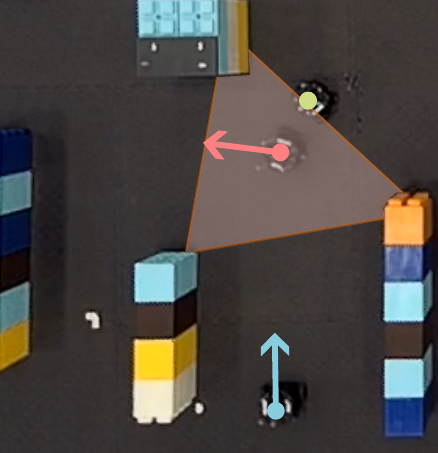}
         \caption{}
         \label{fig:expS3}
     \end{subfigure}
     \begin{subfigure}[b]{.25\linewidth}
         \centering
         \includegraphics[width=\linewidth]{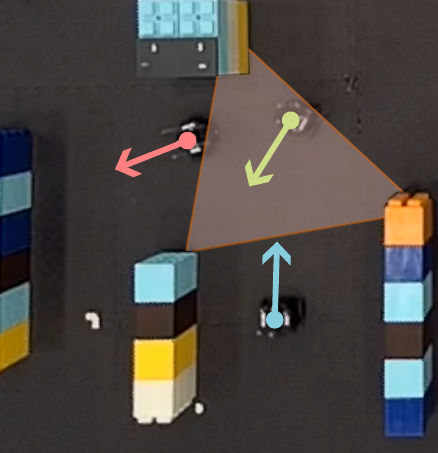}
         \caption{}
         \label{fig:expS4}
     \end{subfigure}
     \begin{subfigure}[b]{.25\linewidth}
         \centering
         \includegraphics[width=\linewidth]{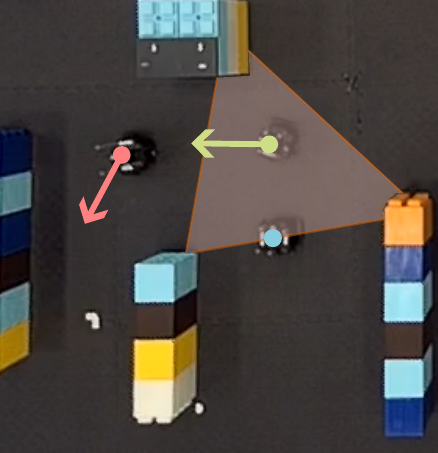}
         \caption{}
         \label{fig:expS6}
     \end{subfigure}
     \begin{subfigure}[b]{.25\linewidth}
         \centering
         \includegraphics[width=\linewidth]{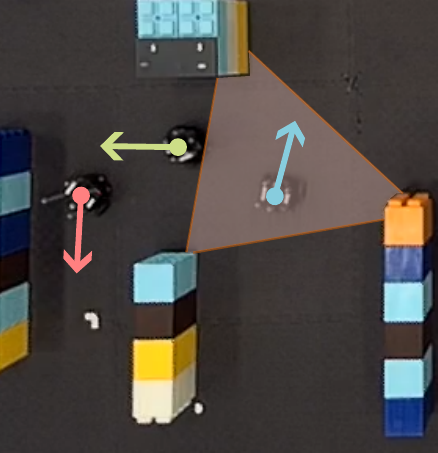}
         \caption{}
         \label{fig:expS6}
     \end{subfigure}
    \caption{Snapshots from an experiment where the robots navigate through one of the semantic regions of a map (an intersection) using the conflict-free paths generated by PM-CBS. The arrows indicate the direction of the robot agents. When a robot is waiting, it is marked with a dot.} 
    \label{fig:expS}
\end{figure}
The proposed method, using the CBS approach (PM-CBS), was then used to find a path for each robot from its current position to a random intersection in the area. The intersections were selected so they were not in the same region as the robot's initial region, and each intersection was assigned to only one robot. Once all robots reached their target regions, a new set of intersections was selected, and a new MAPF (Multi-Agent Pathfinding) problem was solved using PM-CBS. 

This process was repeated for 100 instances, and during the entire experiment, no physical collisions occurred between the robots, thus validating and demonstrating the efficacy of the PM-CBS method. The TurtleBots3 was controlled, and MAPF was solved on a remote computer. The robot had a fixed speed of $R_{speed}$ = 0.15 m/s, and $I_{margin}$ was set to 1.3. The PM-CBS was limited to 30 seconds.

In the experiment, the median computation time was 1.34 ms with an average combined path length of 7.15 m and a success rate of 98.31 \%.
Figure \ref{fig:expS} presents a scenario from the experimental runs that illustrate conflict-free navigation using PM-CBS through one of the map's semantic regions (an intersection). At any point, the intersection is occupied by only one agent, during which other agents wait at the openings of the intersection (enforced through time-restricted path planning presented in section \ref{sec:pathPlan}). The supplementary video accompanying this article highlights more such runs from the experimental validation.

\section{Conclusion}
This article proposed a novel approach to the problem of multi-agent path finding (MAPF). The proposed method merges traditional Conflict-based Search (CBS) with a structural-semantic topometric map. The article introduced generalized definitions of region and opening conflicts, incorporating continuous time constraints and time-restricted path planning, which are more applicable to real-world robot behaviors than classical conflict-based search. The proposed approach overcomes several issues inherent in classical conflict-based search approaches, such as simplistic assumptions about the robots' motion and computational inefficiency. The proposed approach was validated in a benchmarking simulation environment and in an experimental scenario involving nonholonomic Turtlebot3 robots.

Future work involves developing and implementing more advanced collision detection to allow multiple agents to travel in the same region as long as they are not at risk of collision. Additionally, we plan to adapt the method into a real-time planning framework for multiple agents.

\FloatBarrier

\newpage
\bibliographystyle{./IEEEtranBST/IEEEtran}
\bibliography{./IEEEtranBST/IEEEabrv,references}

\begin{thebibliography}{10}
\providecommand{\url}[1]{#1}
\csname url@rmstyle\endcsname
\providecommand{\newblock}{\relax}
\providecommand{\bibinfo}[2]{#2}
\providecommand\BIBentrySTDinterwordspacing{\spaceskip=0pt\relax}
\providecommand\BIBentryALTinterwordstretchfactor{4}
\providecommand\BIBentryALTinterwordspacing{\spaceskip=\fontdimen2\font plus
\BIBentryALTinterwordstretchfactor\fontdimen3\font minus \fontdimen4\font\relax}
\providecommand\BIBforeignlanguage[2]{{%
\expandafter\ifx\csname l@#1\endcsname\relax
\typeout{** WARNING: IEEEtran.bst: No hyphenation pattern has been}%
\typeout{** loaded for the language `#1'. Using the pattern for}%
\typeout{** the default language instead.}%
\else
\language=\csname l@#1\endcsname
\fi
#2}}

\bibitem{stern2019multi}
R.~Stern, N.~Sturtevant, A.~Felner, S.~Koenig, H.~Ma, T.~Walker, J.~Li, D.~Atzmon, L.~Cohen, T.~Kumar, \emph{et~al.}, ``Multi-agent pathfinding: Definitions, variants, and benchmarks,'' in \emph{Proceedings of the International Symposium on Combinatorial Search}, vol.~10, no.~1, 2019, pp. 151--158.

\bibitem{ma2019lifelong}
H.~Ma, W.~H{\"o}nig, T.~S. Kumar, N.~Ayanian, and S.~Koenig, ``Lifelong path planning with kinematic constraints for multi-agent pickup and delivery,'' in \emph{Proceedings of the AAAI Conference on Artificial Intelligence}, vol.~33, no.~01, 2019, pp. 7651--7658.

\bibitem{honig2018trajectory}
W.~H{\"o}nig, J.~A. Preiss, T.~S. Kumar, G.~S. Sukhatme, and N.~Ayanian, ``Trajectory planning for quadrotor swarms,'' \emph{IEEE Transactions on Robotics}, vol.~34, no.~4, pp. 856--869, 2018.

\bibitem{ma2021distributed}
Z.~Ma, Y.~Luo, and H.~Ma, ``Distributed heuristic multi-agent path finding with communication,'' in \emph{2021 IEEE International Conference on Robotics and Automation (ICRA)}.\hskip 1em plus 0.5em minus 0.4em\relax IEEE, 2021, pp. 8699--8705.

\bibitem{lafmejani2021nonlinear}
A.~S. Lafmejani and S.~Berman, ``Nonlinear mpc for collision-free and deadlock-free navigation of multiple nonholonomic mobile robots,'' \emph{Robotics and Autonomous Systems}, vol. 141, p. 103774, 2021.

\bibitem{park2023dlsc}
J.~Park, Y.~Lee, I.~Jang, and H.~J. Kim, ``Dlsc: Distributed multi-agent trajectory planning in maze-like dynamic environments using linear safe corridor,'' \emph{IEEE Transactions on Robotics}, vol.~39, no.~5, pp. 3739--3758, 2023.

\bibitem{Multi_agent_Niklas}
N.~Dahlquist, B.~Lindqvist, A.~Saradagi, and G.~Nikolakopoulos, ``Reactive multi-agent coordination using auction-based task allocation and behavior trees,'' in \emph{2023 IEEE Conference on Control Technology and Applications (CCTA)}, 2023, pp. 829--834.

\bibitem{standley2010finding}
T.~Standley, ``Finding optimal solutions to cooperative pathfinding problems,'' in \emph{Proceedings of the AAAI conference on artificial intelligence}, vol.~24, no.~1, 2010, pp. 173--178.

\bibitem{yu2015optimal}
J.~Yu and S.~M. LaValle, ``Optimal multi-robot path planning on graphs: Structure and computational complexity,'' \emph{arXiv preprint arXiv:1507.03289}, 2015.

\bibitem{ma2019searching}
H.~Ma, D.~Harabor, P.~J. Stuckey, J.~Li, and S.~Koenig, ``Searching with consistent prioritization for multi-agent path finding,'' in \emph{Proceedings of the AAAI conference on artificial intelligence}, vol.~33, no.~01, 2019, pp. 7643--7650.

\bibitem{wagner2015subdimensional}
G.~Wagner and H.~Choset, ``Subdimensional expansion for multirobot path planning,'' \emph{Artificial intelligence}, vol. 219, pp. 1--24, 2015.

\bibitem{sharon2015conflict}
G.~Sharon, R.~Stern, A.~Felner, and N.~R. Sturtevant, ``Conflict-based search for optimal multi-agent pathfinding,'' \emph{Artificial intelligence}, vol. 219, pp. 40--66, 2015.

\bibitem{boyarski2015icbs}
E.~Boyarski, A.~Felner, R.~Stern, G.~Sharon, O.~Betzalel, D.~Tolpin, and E.~Shimony, ``Icbs: The improved conflict-based search algorithm for multi-agent pathfinding,'' in \emph{Proceedings of the International Symposium on Combinatorial Search}, vol.~6, no.~1, 2015, pp. 223--225.

\bibitem{gange2019lazy}
G.~Gange, D.~Harabor, and P.~J. Stuckey, ``Lazy cbs: implicit conflict-based search using lazy clause generation,'' in \emph{Proceedings of the international conference on automated planning and scheduling}, vol.~29, 2019, pp. 155--162.

\bibitem{via2020efficient}
N.-h. M.~R. via Prioritized, ``Efficient trajectory planning for multiple non-holonomic mobile robots via prioritized trajectory optimization,'' 2020.

\bibitem{tajbakhsh2024conflict}
A.~Tajbakhsh, L.~T. Biegler, and A.~M. Johnson, ``Conflict-based model predictive control for scalable multi-robot motion planning,'' in \emph{2024 IEEE International Conference on Robotics and Automation (ICRA)}.\hskip 1em plus 0.5em minus 0.4em\relax IEEE, 2024, pp. 14\,562--14\,568.

\bibitem{mcbeth2023scalable}
C.~McBeth, J.~Motes, D.~Uwacu, M.~Morales, and N.~M. Amato, ``Scalable multi-robot motion planning for congested environments with topological guidance,'' \emph{IEEE Robotics and Automation Letters}, 2023.

\bibitem{li2019symmetry}
J.~Li, D.~Harabor, P.~J. Stuckey, H.~Ma, and S.~Koenig, ``Symmetry-breaking constraints for grid-based multi-agent path finding,'' in \emph{Proceedings of the AAAI conference on artificial intelligence}, vol.~33, no.~01, 2019, pp. 6087--6095.

\bibitem{li2021pairwise}
J.~Li, D.~Harabor, P.~J. Stuckey, H.~Ma, G.~Gange, and S.~Koenig, ``Pairwise symmetry reasoning for multi-agent path finding search,'' \emph{Artificial Intelligence}, vol. 301, p. 103574, 2021.

\bibitem{Remolina2004}
E.~Remolina and B.~Kuipers, ``{Towards a general theory of topological maps},'' vol. 152, no.~1, pp. 47--104.

\bibitem{Kostavelis2015}
I.~Kostavelis and A.~Gasteratos, ``{Semantic mapping for mobile robotics tasks: A survey},'' vol.~66, pp. 86--103.

\bibitem{FREDRIKSSON_SEMANTIC_MAPPING}
S.~Fredriksson, A.~Saradagi, and G.~Nikolakopoulos, ``Semantic and topological mapping using intersection identification,'' \emph{IFAC-PapersOnLine}, vol.~56, no.~2, pp. 9251--9256, 2023, 22nd IFAC World Congress.

\bibitem{barer2014suboptimal}
M.~Barer, G.~Sharon, R.~Stern, and A.~Felner, ``Suboptimal variants of the conflict-based search algorithm for the multi-agent pathfinding problem,'' in \emph{Proceedings of the international symposium on combinatorial Search}, vol.~5, no.~1, 2014, pp. 19--27.

\bibitem{Hector2011}
S.~Kohlbrecher, J.~Meyer, O.~von Stryk, and U.~Klingauf, ``A flexible and scalable slam system with full 3d motion estimation,'' in \emph{Proc. IEEE International Symposium on Safety, Security and Rescue Robotics (SSRR)}.\hskip 1em plus 0.5em minus 0.4em\relax IEEE, November 2011.

\end{thebibliography}
\end{document}